\documentclass[sigconf, nonacm]{acmart}
\renewcommand\footnotetextcopyrightpermission[1]{}
\AtBeginDocument{%
  }

\usepackage{balance}
\usepackage{romannum}
\usepackage{float}
\usepackage{amsmath}
\usepackage{subcaption}  
\usepackage{algorithm}
\usepackage{algpseudocode} 
\usepackage{algpseudocode}
\usepackage{float} 

\begin{document}
\raggedbottom
\title{DroneVLA: VLA based Aerial Manipulation}



\author{Fawad Mehboob}
\email{Fawad.Mehboob@skoltech.ru}
\authornote{Authors contributed equally to this research.}
\affiliation{%
  \institution{Skoltech}
  \city{Moscow}
  \country{Russia}
}

\author{Monijesu James}
\email{Monijesu.James@skoltech.ru}
\authornotemark[1]
\affiliation{%
  \institution{Skoltech}
  \city{Moscow}
  \country{Russia}
}

\author{Amir Habel}
\email{Amir.Habel@skoltech.ru}
\authornotemark[1]
\affiliation{%
  \institution{Skoltech}
  \city{Moscow}
  \country{Russia}
}

\author{Jeffrin Sam}
\email{Jeffrin.Sam@skoltech.ru}
\affiliation{%
  \institution{Skoltech}
  \city{Moscow}
  \country{Russia}
}

\author{Miguel Altamirano Cabrera}
\email{m.altamirano@skoltech.ru}
\affiliation{%
  \institution{Skoltech}
  \city{Moscow}
  \country{Russia}
}

\author{Dzmitry Tsetserukou}
\email{d.tsetserukou@skoltech.ru}
\affiliation{%
  \institution{Skoltech}
  \city{Moscow}
  \country{Russia}
}

\renewcommand{\shortauthors}{Mehboob et al.}

\begin{abstract}
As aerial platforms evolve from passive observers to active manipulators, the challenge shifts toward designing intuitive interfaces that allow non-expert users to command these systems naturally. This work introduces a novel concept of autonomous aerial manipulation system capable of interpreting high-level natural language commands to retrieve objects and deliver them to a human user. The system is intended to integrate a MediaPipe based on Grounding DINO and a Vision-Language-Action (VLA) model with a custom-built drone equipped with a 1-DOF gripper and an Intel RealSense RGB-D camera. VLA performs semantic reasoning to interpret the intent of a user prompt and generates a prioritized task queue for grasping of relevant objects in the scene.
Grounding DINO and dynamic A* planning algorithm are used to navigate and safely relocate the object. To ensure safe and natural interaction during the handover phase, the system employs a human-centric controller driven by MediaPipe. This module provides real-time human pose and orientation estimation, allowing the drone to employ visual servoing to maintain a stable, distinct position directly in front of the user, facilitating a comfortable handover. We demonstrate the system’s efficacy through real-world experiments for localization and navigation, highlighting the feasibility of VLA for aerial manipulation operations. 
\end{abstract}

\ccsdesc[500]{Human-centered computing~Human–robot interaction}
\ccsdesc[300]{Computing methodologies~Computer vision~Scene understanding}
\ccsdesc[100]{Computing methodologies~Artificial intelligence~Natural language interfaces}
\ccsdesc[100]{Computer systems organization~Embedded and cyber-physical systems~Robotics}

\keywords{Aerial Manipulation, Vision-Language-Action Models, Human-Robot Interaction, Visual Surveying, Robotic Fetch-and-Carry}



\maketitle

\section{Introduction}
Unmanned Aerial Vehicles (UAVs) have evolved to be used in an increasing number of applications, ranging from surveillance and inspection to agriculture, search and rescue missions, and much more. More recently, significant interest has developed in indoor UAV operations, such as inspection in industrial environments and warehouses. This has paved a way for a
a growing interest in integrating robotic grippers with the UAVs 
enabling them to manipulate objects.
As a result, an emerging field of aerial manipulation has captured the interest of UAV researchers, transforming UAVs into Unmanned Aerial Manipulators (UAMs)
\begin{figure}[t!]
    \centering
    \includegraphics[width=0.8\linewidth]{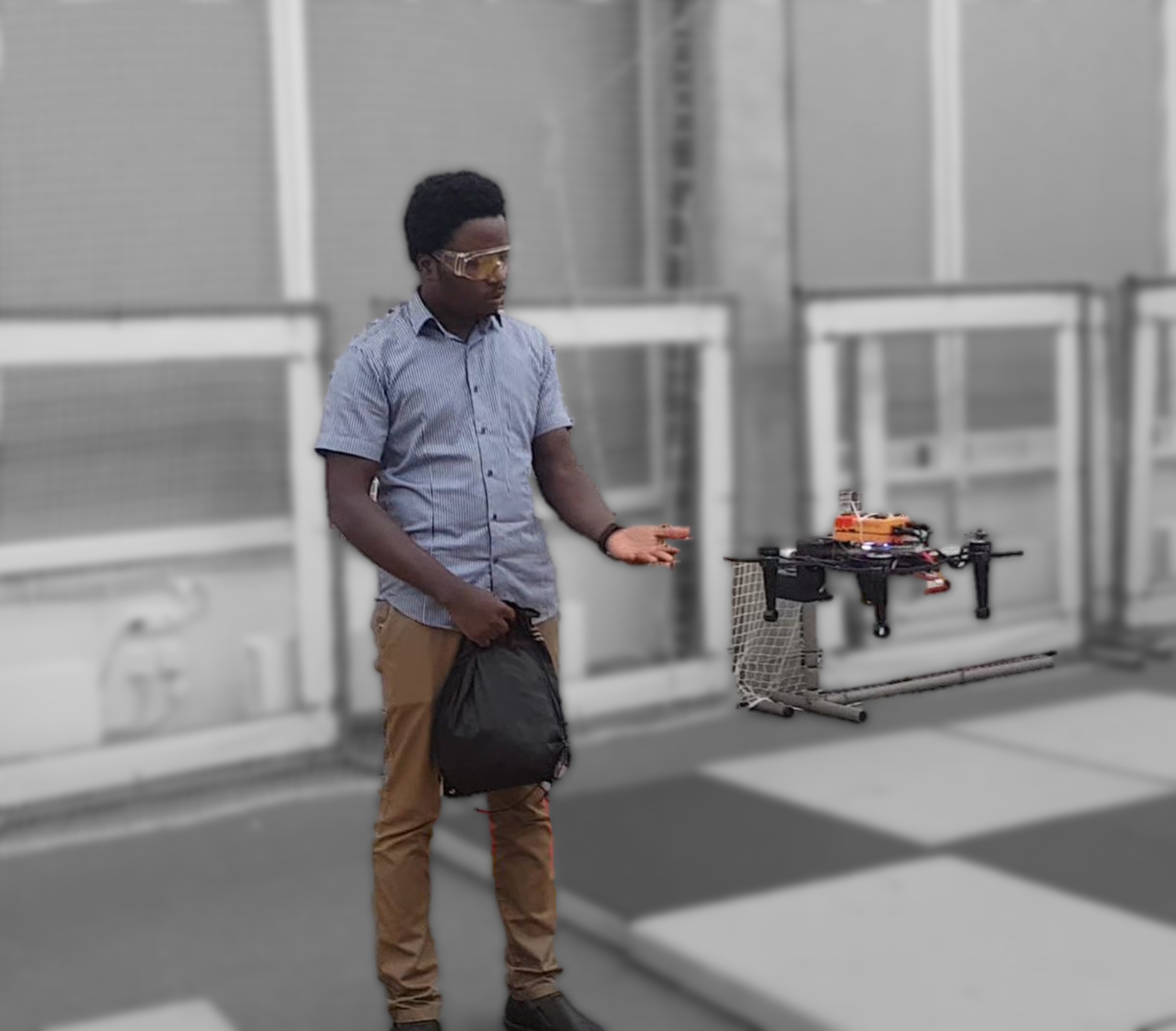}
    \caption{Human Localization and Navigation by drone}
    \label{fig:Human_drone_interaction}
\end{figure}

Combining the high motion agility of UAVs with the manipulability of robotic arms is an interesting yet challenging field of research. A very useful scenario of indoor UAV use is the ability to open doors; this problem has been studied in \cite{tsukagoshi2015aerial} by proposing a novel force exertion scheme. Similar force exertion scenarios are studied for valve-turning operations using thrust vectoring in \cite{zhao2022forceful}. Moreover, an interesting operation of UAMs is demonstrated in aerial calligraphy, as proposed in \cite{guo2024flying}. In addition to this, the most commonly studied use case is the transportation of objects using UAMs \cite{khamseh2018aerial},  which also involves dealing with external forces, changes in vehicle inertia, and stability issues. A comprehensive review of UAMs discussing their utility and realization challenges is provided in \cite{khamseh2018aerial} and \cite{ruggiero2018aerial}. More recently, the focus of aerial manipulation has shifted to interaction and collaboration with humans. To this end, a master-slave tele-operation framework has been demonstrated on an aerial manipulator in \cite{meng2025bilateral}.

Additionally, the idea of incorporating Large Language Models (LLMs) to take prompts from humans and perform tasks in aerial manipulation seems promising. However, the reliability of such a framework is a critical challenge, especially in environments where drones interact with humans and the safety of both the human and the environment is of high concern. To this end a Vision-Language Model (VLM)-based mission was proposed in \cite{mishra2025aermani}, with considerable efforts to make prompt-based aerial manipulation reliable by mitigating hallucinations using structured prompting and a predefined skill library. However, this approach lacks authority over drone control, which is critical in such scenarios given the highly dynamic nature of drone flights. Quadrotor flights are prone to high disturbances, especially close to the ground and near physical objects due to downwash produced by the propellers while hovering \cite{morishita2024downwash}. To address this issue, a more robust framework is required that accounts for high-level drone control using VLA similar to as demonstrated for UAV flights in \cite{sautenkov2025uav,serpiva2025racevla}. Therefore, in this study, we propose a conceptual Vision-Language-Action (VLA) model for aerial manipulation of objects in a human-robot interaction environment. Our VLA is implemented on a quadrotor UAV equipped with a simple gripping mechanism, designed to pick up and place objects from locations such as tables and shelves. To the best of our knowledge, this is the first proposal of VLA for aerial manipulation tasks.

\section{System Architecture}

The system comprises four major subsystems: (1) hardware platform, (2) VLA-based semantic reasoning module, (3) perception pipeline with open-vocabulary detection, and (4) human-centric handover controller. Fig.~\ref{fig:Architecture} illustrates the overall system architecture.

\subsection{Hardware Platform}
The hardware consists of a simple rigid-body open-close gripping mechanism powered by a Dynamixel AX-12A motor mounted horizontally at the front of a quadrotor UAV.
Considerable care has been taken in positioning the gripper away from the drone body to avoid interference while picking up objects, this helps mitigate the downwash which effects flight stability and also cause a problem of blowing the objects away from the drone while grasping. However, a trade-off must be made between the stability of the drone and the distance of the gripper from the drone's center of mass (COM).
\begin{figure}[t!]
    \centering
    \includegraphics[width=1.0\linewidth]{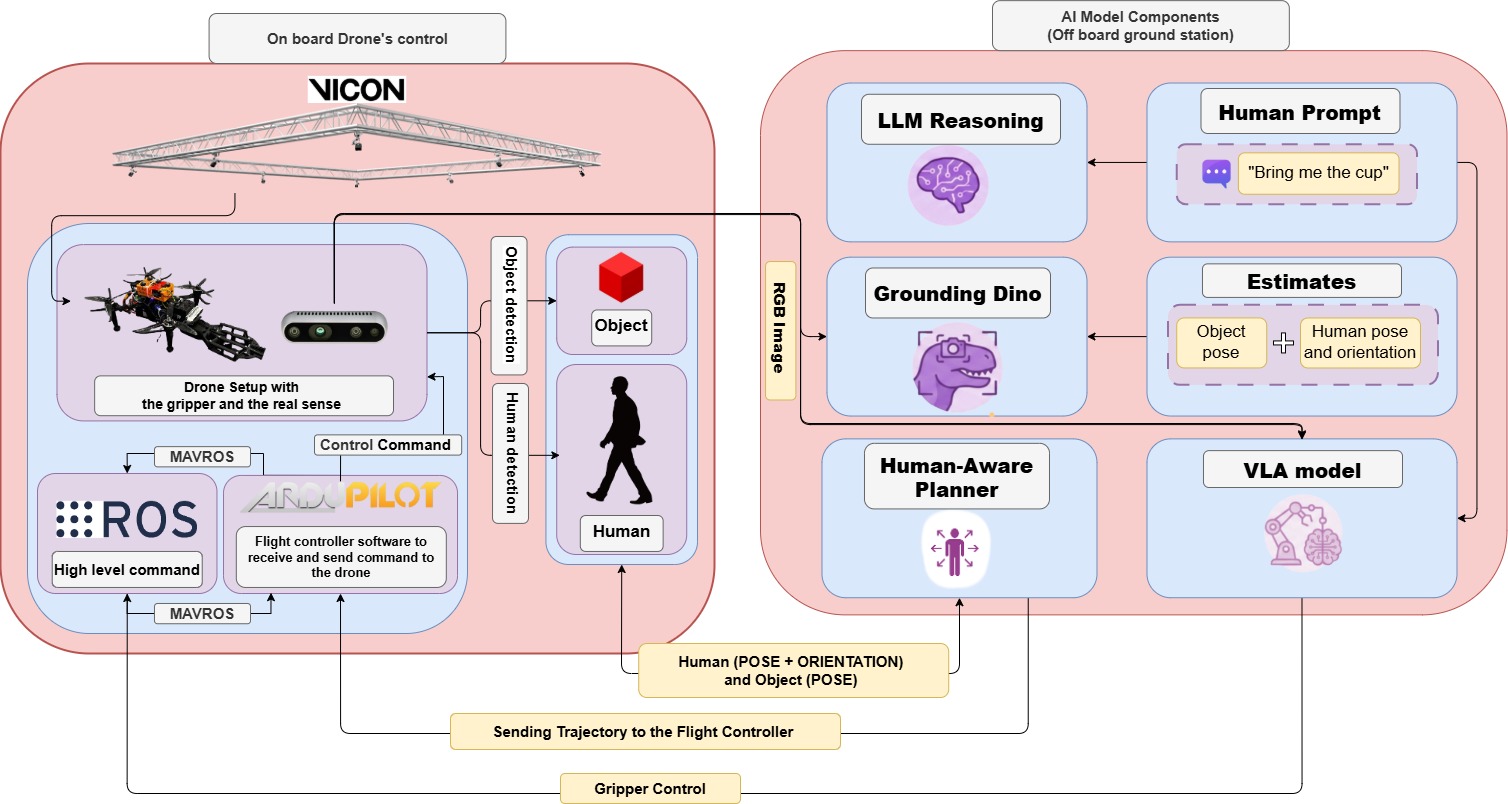}
    \caption{Overview of the VLA based aerial manipulation and Path Planning Architecture}
    \label{fig:Architecture}
\end{figure}

All modules communicate via ROS2 topics using standard message types. The distributed architecture enables computationally intensive perception and reasoning tasks to run offboard while maintaining low-latency control loops onboard the OrangePi Fig.~\ref{fig:Architecture}.

\subsection{Vision-Language-Action (VLA) Module}
To enable intuitive control on resource-constrained hardware, we employ a lightweight Vision-Language-Action (VLA) model inspired by architectures such as \textbf{TinyVLA} \cite{wen2025tinyvla}, which aims to bring multi-modal capabilities to embedded hardware. Here we propose a conceptual VLA system validated in Unity Simulation for detecting objects and grasping using a gripper.


The proposed VLA module accepts two primary inputs:

\begin{itemize}
\item \textbf{Visual Observation ($I_t$):} A time-indexed RGB image stream captured from the onboard RealSense camera. Prior to fusion, the visual stream is down-sampled to $224 \times 224$ resolution and encoded using a compact convolutional visual backbone.
\item \textbf{Language Prompt ($L$):} A natural language instruction from the operator specifying an object and intended action (e.g., 'pick up the red screwdriver'). In the current prototype, text is supplied via keyboard input.
\end{itemize}

The VLA architecture is deployed on the ground station rather than onboard the UAV, enabling increased model complexity without incurring latency or power constraints on the aerial platform. The TinyVLA is composed of a 6-layer ViT-Tiny visual encoder (patch size 16×16, 192-dim latent) and a 4-layer text transformer (256-dim hidden, 8 attention heads) for natural language prompt processing. Visual frames streamed from the RealSense camera over Wi-Fi are downsampled to 224×224 resolution and processed asynchronously at 15–20 Hz. The fused latent representation is passed through a 2-layer MLP action policy head that predicts discrete gripper actions {Open,Close}. The resulting model executes with sub-5 ms inference on a desktop-class RTX GPU. The ground station deployment also allows future expansion of the action vocabulary, multi-object grounding, and affordance reasoning without modifying the UAV flight stack.

Given the single-degree-of-freedom parallel gripper mounted beneath the quadrotor, the output action space is defined as a binary actuator command:

\begin{equation}
a_t = \pi(I_t, L) \in {\text{Open},\ \text{Close}}
\end{equation}

As evident from the action-space, the VLA implements a simple opening and closing action upon detection of being in the vicinity of the object of interest. Meanwhile, spatial positioning and approach trajectory are handled by a  human-centric controller driven by MediaPipe as illustrated in Fig.~\ref{fig:Architecture} and explained in the subsequent sections. This architecture decouples semantic manipulation reasoning from UAV stabilization and navigation, and serves as a first step that can be extended to a more general end-to-end architecture to apply a full 5-DOF control, that includes velocities in the x,y,z coordinates,yaw angular rate, and gripper state.

This direct mapping from perception to actuation allows the VLA to focus exclusively on grasping logic. The VLA continuously infers the correct gripper state based on the visual proximity to the object described in the text prompt, while the drone's position controller handles the approach trajectory independently.

\subsection{Perception Pipeline: Open-Vocabulary Detection and 3D Localization}
Traditional object detection systems typically require retraining for new object categories. To enable true open-vocabulary manipulation, we integrate \textbf{Grounding DINO} \cite{liu2023grounding}, a transformer-based open-set object detector that performs cross-modal fusion of vision and language features. Grounding DINO takes as input:
\textbf{RGB images} from RealSense cameras, and \textbf{text prompts} specifying target objects(e.g., "red cup", "person in red shirt", "screwdriver"). 

The detector outputs bounding boxes with confidence scores for all instances that match the text query. This enables zero-shot detection of arbitrary objects without task-specific training.

\textbf{3D Localization:} For each bounding boxes detected by Grounding DINO, we compute the centroid of bounding box in image coordinates \((u,v)\), query the aligned depth map at \((u,v)\) to obtain depth \(d\), deproject to 3D camera coordinates and transform from camera frame to world frame using ROS2 TF2 transforms.\\
This pipeline provides real-time 3D positions of the target objects and humans in the global reference frame, enabling subsequent visual servoing control.
\subsection{Human-Centric Handover Controller}
\begin{figure}[H]
    \centering
    \includegraphics[width=1.0\linewidth]{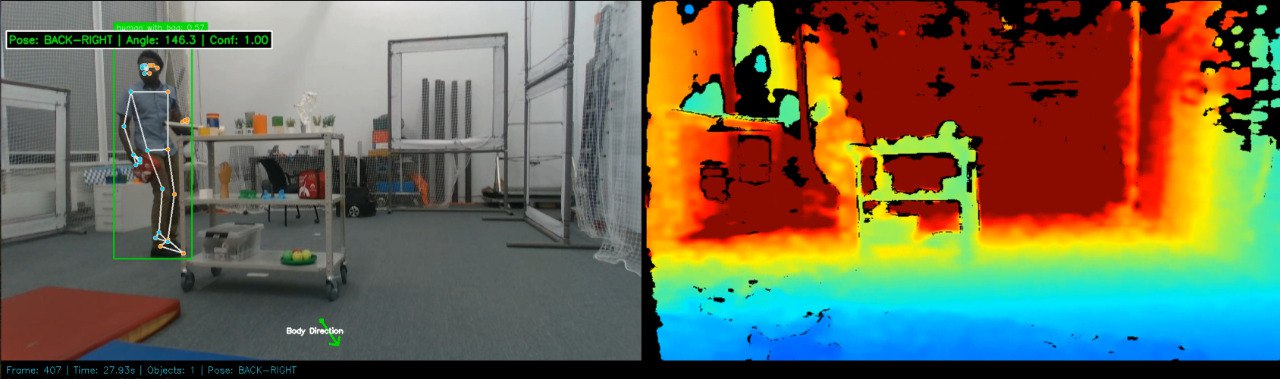}
    \caption{Grounding DINO bounding box object detection and humans in RGB images from drone cameras}
    \label{fig:placeholder}
\end{figure}

Safe and natural human-robot handover requires understanding the human's pose, orientation, and preferred interaction geometry. We employ \textbf{MediaPipe Pose Landmarker} to estimate human pose in real-time.\\ MediaPipe outputs: 33 skeletal keypoints in both normalized 2D image coordinates and 3D world coordinates, body orientation computed from shoulder and hip joint vectors, and Hand position estimate to determine handover side preference as shown in Fig.~\ref{fig:placeholder} .

\textbf{Human Orientation Estimation: }
Let \(\mathbf{p}_{\text{left\_shoulder}}\) and \(\mathbf{p}_{\text{right\_shoulder}}\) denote the 3D positions of the left and right shoulder landmarks, the arc-tangent of the difference of these landmarks is used to compute the torso forward direction (yaw angle):

\begin{equation}
\mathbf{v}_{\text{torso}} = \mathbf{p}_{\text{right\_shoulder}} - \mathbf{p}_{\text{left\_shoulder}},
\end{equation}

\begin{equation}
\theta_{\text{yaw}} = \text{arctan}(\mathbf{v}_{\text{torso},y}, \mathbf{v}_{\text{torso},x})
\end{equation}

The perpendicular direction to the torso indicates the person's facing direction.

\textbf{Handover Pose Computation:}
Based on HRI handover literature \cite{strabala2013toward,nemlekar2019object}, we compute a target handover pose in form of Distance (approximately 0.6-0.8, in from of the human), Height (Chest level at approximately 1.0-1.3m from ground), and orientation (Drone faces human directly, aligned with human's facing direction)

The target handover pose in the world frame is:
\begin{equation}
{}^W\mathbf{p}_{\text{handover}} = {}^W\mathbf{p}_{\text{human}} + d_{\text{handover}} \cdot \begin{bmatrix} \cos(\theta_{\text{yaw}} + \pi) \\ \sin(\theta_{\text{yaw}} + \pi) \\ 0 \end{bmatrix} + \begin{bmatrix} 0 \\ 0 \\ h_{\text{chest}} \end{bmatrix},
\end{equation}
where \(d_{\text{handover}}\) is the desired standoff distance and \(h_{\text{chest}}\) is the vertical offset to chest height.

\textbf{Visual Servoing Control}:
We implement a hybrid servoing approach combining \textbf{position-based visual servoing (PBVS)} and \textbf{image-based visual servoing (IBVS)}.

The \textbf{PBVS} is used for coarse navigation to the object and handover locations using 3D pose estimates from depth cameras. 
    
While the \textbf{IBVS} is used for fine alignment during grasping and handover, directly using 2D image features to minimize drift and maintain object/human visibility. 

The image-based error \(\mathbf{e}_{\text{img}} = \mathbf{s} - \mathbf{s}^*\) (current vs. desired feature positions) drives the control.

Safety checks and failure recovery behaviors (e.g., lost object detection, communication timeout) are integrated at each state transition.

\subsection{Motion planning with human awareness}
\label{subsec:motion_planning}

The global motion planner operates in the horizontal plane using a
grid–based A* search. The room
$[x_{\min},x_{\max}] \times [y_{\min},y_{\max}]$ is discretized into a
uniform occupancy grid with resolution $\Delta$, where each cell
represents a candidate drone position and the occupancy value
$\mathit{occ}(i,j) \in \{0,1\}$ encodes free space or collision.
Circular obstacles are rasterized by marking all cells whose centers lie
within the corresponding radius; in particular, the human is modeled as
a cylindrical obstacle with radius equal to the physical body radius
plus an additional safety margin, enforcing a minimum stand-off distance
throughout the mission. For each leg of the task (home–object,
object–human, human–home), continuous start and goal positions are
projected to the nearest free cells and A* is run on the 8-connected
grid with unit cost for axial moves and $\sqrt{2}$ for diagonal moves,
and a Euclidean heuristic in grid index space. The resulting sequence of
cells is converted back to world coordinates and passed through a
line-of-sight smoothing step, which
removes redundant waypoints while preserving collision-free straight
segments. The final polyline is then sent to the low-level controller as
a sequence of position setpoints with altitude specified by the mission
logic.

\section{System Evaluation}
\subsection{Experimental Procedure}

\textbf{Setup.}
Experiments were conducted in an indoor laboratory space of approximately
$6~\mathrm{m} \times 6~\mathrm{m}$ equipped with an overhead Vicon motion
capture system, which provided ground–truth pose for the drone, the table,
and the human participant. A rectangular table was placed in the center of
the workspace and populated with $15$–$20$ everyday objects (cups, tools,
plants etc) arranged in different configurations. A human stood at a
designated handover location approximately $2~\mathrm{m}$ from the table.
The onboard perception stack (Grounding DINO, depth sensing, and
MediaPipe) and the human–aware A* motion planner were evaluated jointly as
an integrated pipeline for scene surveying, object and human localization,
and safe navigation.

\textbf{Trials.}
We conducted \textbf{$N = 10$} trials with randomized object
arrangements. Each trial consisted of:
\begin{enumerate}
    \item Autonomous take-off and survey of the workspace;
    \item Object and human identification, classification, and localization;
    \item Motion planning and execution from home to object, object grasp,
          and human-aware navigation from object to human;
    \item Logging of trial outcome (success/failure), execution time, and
          failure mode (if any).
\end{enumerate}

All onboard sensor data, perception outputs, and control commands were
recorded as ROS~2 bag files for offline analysis. Vicon trajectories were
time-synchronized with the drone’s onboard logs to enable quantitative
evaluation of perception accuracy and motion-planning performance.

\subsection{Discussion}
In the experiments performed, we demonstrated a fully functioning, end-to-end MediaPipe based on grounding DINO for object and human identification and localization.

Recent work such as RaceVLA \cite{serpiva2025racevla} demonstrate the power of end-to-end learning, where a single transformer model maps raw visual inputs directly to 4D control commands. While RaceVLA
achieves impressive agility and human-like behaviors in high speed navigation tasks, it operates as a "black box." In the context to aerial manipulation, this poses a safety risk: hallucination in the VLA could result in a direct collision command. Our approach mitigates this by decoupling reasoning from actuation. By using VLA solely for semantic intent and relying on deterministic visual servoing for the \textit{how}, DroneVLA ensures drone's motion is governed by verifiable control laws.

On the other end of the spectrum, systems like AERMANI-VLM used structured prompting to generate high-level reasoning traces that select discrete skill primitives. While robust for planning, these systems can suffer from rigidity in dynamic execution. DroneVLA occupies a middle ground; it employs the semantic flexibility of VLMs for object grounding but integrates tight closed-loop visual servoing to handle the continuous adjustments required for grasping.
\begin{figure}[t]
    \centering
    \includegraphics[width=1.0\linewidth]{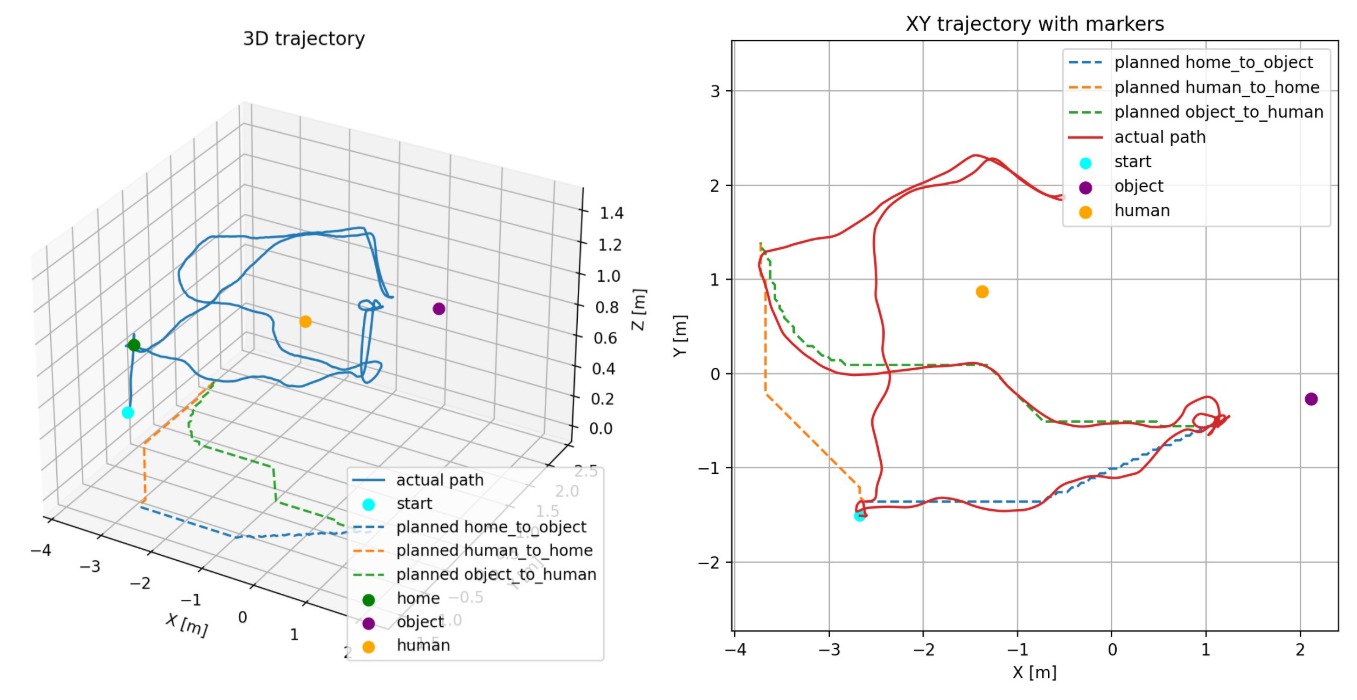}
    \caption{2D and 3D views of the human–aware A* motion plan,
             showing the reference trajectory, planned path, safety
             margins around the human and table with achieved trajectory in real flight.}
    \label{fig:planning}
\end{figure}

A total of 10 experiments were performed and all experiments were successfully implemented with a max error of 0.164m, a mean euclidean error of 0.070m, and a root mean squared error of 0.084m. This can be visualized in the home-to-object and object-to-human trajectories of the 2-3-D plot of Fig.~\ref{fig:planning}. Also it is noticeable that the drone stays a safe distance of 1 m away from the human achieving the desired safe handover trajectory.

Although the VLA architecture was not implemented in the real-world flight control loop, a Unity-based simulation was performed to strictly validate the grasping policy. 10 simulation experiments were conducted where visual frames were fed to the model to verify the correctness of the predicted gripper action. In all trials, the model correctly identified the appropriate binary action (open/close) based on the image state. It is important to note that the validated VLA operates as a binary actuation model: it analyzes the received image state—whether from simulation or real-world feeds—and triggers a discrete `open' or `close' gripper command. This confirms the feasibility of the vision-to-action mapping and validates the model's decision-making logic prior to physical deployment.

\section{Conclusion and Future Work}
Having established an end-to-end MediaPipe and Grounding DINO framework for localization in versatile environments, we proposed and validated a novel concept for VLA-based aerial manipulation. The results demonstrate the robustness of the perception stack in real-world flights and the accuracy of the VLA's grasp validation logic, which was successfully verified on both simulated and real-world visual data. This work serves as a foundational step towards embodied aerial intelligence.

Future work will focus on transitioning from open-loop validation to closed-loop real-time control. Crucially, we intend to leverage this complete framework as a data collection pipeline; by recording the visual inputs and corresponding actions during these flights, we will generate a comprehensive dataset to train a custom, end-to-end VLA model for full 5-DOF control. Finally, we aim to extend this architecture to industrial domains by developing a specialist VLA model tailored for complex indoor navigation and manipulation tasks, enabling fully autonomous workflows in dynamic operational environments.


\section*{Acknowledgements} 
Research reported in this publication was financially supported by the RSF grant No. 24-41-02039.



\begin{thebibliography}{14}


\ifx \showCODEN    \undefined \def \showCODEN     #1{\unskip}     \fi
\ifx \showISBNx    \undefined \def \showISBNx     #1{\unskip}     \fi
\ifx \showISBNxiii \undefined \def \showISBNxiii  #1{\unskip}     \fi
\ifx \showISSN     \undefined \def \showISSN      #1{\unskip}     \fi
\ifx \showLCCN     \undefined \def \showLCCN      #1{\unskip}     \fi
\ifx \shownote     \undefined \def \shownote      #1{#1}          \fi
\ifx \showarticletitle \undefined \def \showarticletitle #1{#1}   \fi
\ifx \showURL      \undefined \def \showURL       {\relax}        \fi
\providecommand\bibfield[2]{#2}
\providecommand\bibinfo[2]{#2}
\providecommand\natexlab[1]{#1}
\providecommand\showeprint[2][]{arXiv:#2}

\bibitem[Guo et~al\mbox{.}(2024)]%
        {guo2024flying}
\bibfield{author}{\bibinfo{person}{X. Guo}, \bibinfo{person}{G. He}, \bibinfo{person}{J. Xu}, \bibinfo{person}{M. Mousaei}, \bibinfo{person}{J. Geng}, \bibinfo{person}{S. Scherer}, {and} \bibinfo{person}{G. Shi}.} \bibinfo{year}{2024}\natexlab{}.
\newblock \showarticletitle{Flying calligrapher: Contact-aware motion and force planning and control for aerial manipulation}.
\newblock \bibinfo{journal}{\emph{IEEE Robotics and Automation Letters}} (\bibinfo{year}{2024}).
\newblock
\href{https://doi.org/10.1109/LRA.2024.3486236}{doi:\nolinkurl{10.1109/LRA.2024.3486236}}


\bibitem[Khamseh et~al\mbox{.}(2018)]%
        {khamseh2018aerial}
\bibfield{author}{\bibinfo{person}{H.B. Khamseh}, \bibinfo{person}{F. Janabi-Sharifi}, {and} \bibinfo{person}{A. Abdessameud}.} \bibinfo{year}{2018}\natexlab{}.
\newblock \showarticletitle{Aerial manipulation—A literature survey}.
\newblock \bibinfo{journal}{\emph{Robotics and Autonomous Systems}}  \bibinfo{volume}{107} (\bibinfo{year}{2018}), \bibinfo{pages}{221--235}.
\newblock
\href{https://doi.org/10.1016/j.robot.2018.03.007}{doi:\nolinkurl{10.1016/j.robot.2018.03.007}}


\bibitem[Liu et~al\mbox{.}(2024)]%
        {liu2023grounding}
\bibfield{author}{\bibinfo{person}{Shilong Liu}, \bibinfo{person}{Zhaoyang Zeng}, \bibinfo{person}{Tianhe Ren}, \bibinfo{person}{Feng Li}, \bibinfo{person}{Hao Zhang}, \bibinfo{person}{Jie Yang}, \bibinfo{person}{Qing Jiang}, \bibinfo{person}{Chunyuan Li}, \bibinfo{person}{Jianwei Yang}, \bibinfo{person}{Hang Su}, \bibinfo{person}{Jun Zhu}, {and} \bibinfo{person}{Lei Zhang}.} \bibinfo{year}{2024}\natexlab{}.
\newblock \bibinfo{title}{Grounding DINO: Marrying DINO with Grounded Pre-Training for Open-Set Object Detection}.
\newblock
\showeprint[arxiv]{2303.05499.}
\newblock
\shownote{Retrieved from https://arxiv.org/abs/2303.05499}.


\bibitem[Meng et~al\mbox{.}(2025)]%
        {meng2025bilateral}
\bibfield{author}{\bibinfo{person}{L. Meng}, \bibinfo{person}{Y. Rong}, {and} \bibinfo{person}{W. Chou}.} \bibinfo{year}{2025}\natexlab{}.
\newblock \showarticletitle{Bilateral Teleoperation of Aerial Manipulator with Hybrid Mapping Framework for Physical Interaction}.
\newblock \bibinfo{journal}{\emph{Sensors}} \bibinfo{volume}{25}, \bibinfo{number}{18} (\bibinfo{year}{2025}), \bibinfo{pages}{5844}.
\newblock
\href{https://doi.org/10.3390/s25185844}{doi:\nolinkurl{10.3390/s25185844}}


\bibitem[Mishra et~al\mbox{.}(2025)]%
        {mishra2025aermani}
\bibfield{author}{\bibinfo{person}{Sarthak Mishra}, \bibinfo{person}{Rishabh~Dev Yadav}, \bibinfo{person}{Avirup Das}, \bibinfo{person}{Saksham Gupta}, \bibinfo{person}{Wei Pan}, {and} \bibinfo{person}{Spandan Roy}.} \bibinfo{year}{2025}\natexlab{}.
\newblock \bibinfo{title}{AERMANI-VLM: Structured Prompting and Reasoning for Aerial Manipulation with Vision Language Models}.
\newblock
\showeprint[arxiv]{2511.01472.}
\newblock
\shownote{Retrieved from https://arxiv.org/abs/2511.01472}.


\bibitem[Morishita et~al\mbox{.}(2024)]%
        {morishita2024downwash}
\bibfield{author}{\bibinfo{person}{Ryosuke Morishita}, \bibinfo{person}{Shin Kawai}, {and} \bibinfo{person}{Hajime Nobuhara}.} \bibinfo{year}{2024}\natexlab{}.
\newblock \showarticletitle{Downwash reduction drone with adaptive rotors and its 3D aerodynamic analysis and stabilization control}.
\newblock \bibinfo{journal}{\emph{IEEE Access}}  \bibinfo{volume}{12} (\bibinfo{year}{2024}), \bibinfo{pages}{22832--22840}.
\newblock
\href{https://doi.org/10.1109/ACCESS.2024.3362639}{doi:\nolinkurl{10.1109/ACCESS.2024.3362639}}


\bibitem[Nemlekar et~al\mbox{.}(2019)]%
        {nemlekar2019object}
\bibfield{author}{\bibinfo{person}{Heramb Nemlekar}, \bibinfo{person}{Dharini Dutia}, {and} \bibinfo{person}{Zhi Li}.} \bibinfo{year}{2019}\natexlab{}.
\newblock \showarticletitle{Object transfer point estimation for fluent human-robot handovers}. In \bibinfo{booktitle}{\emph{Proc. Int. Conf. on Robotics and Automation (ICRA)}}. IEEE, \bibinfo{pages}{2627--2633}.
\newblock
\href{https://doi.org/10.1109/ICRA.2019.8794008}{doi:\nolinkurl{10.1109/ICRA.2019.8794008}}


\bibitem[Ruggiero et~al\mbox{.}(2018)]%
        {ruggiero2018aerial}
\bibfield{author}{\bibinfo{person}{F. Ruggiero}, \bibinfo{person}{V. Lippiello}, {and} \bibinfo{person}{A. Ollero}.} \bibinfo{year}{2018}\natexlab{}.
\newblock \showarticletitle{Aerial manipulation: A literature review}.
\newblock \bibinfo{journal}{\emph{IEEE Robotics and Automation Letters}} \bibinfo{volume}{3}, \bibinfo{number}{3} (\bibinfo{year}{2018}), \bibinfo{pages}{1957--1964}.
\newblock
\href{https://doi.org/10.1109/LRA.2018.2822650}{doi:\nolinkurl{10.1109/LRA.2018.2822650}}


\bibitem[Sautenkov et~al\mbox{.}(2025)]%
        {sautenkov2025uav}
\bibfield{author}{\bibinfo{person}{Oleg Sautenkov}, \bibinfo{person}{Yasheerah Yaqoot}, \bibinfo{person}{Artem Lykov}, \bibinfo{person}{Muhammad~Ahsan Mustafa}, \bibinfo{person}{Grik Tadevosyan}, \bibinfo{person}{Aibek Akhmetkazy}, \bibinfo{person}{Miguel~Altamirano Cabrera}, \bibinfo{person}{Mikhail Martynov}, \bibinfo{person}{Sausar Karaf}, {and} \bibinfo{person}{Dzmitry Tsetserukou}.} \bibinfo{year}{2025}\natexlab{}.
\newblock \showarticletitle{UAV-VLA: Vision-language-action system for large scale aerial mission generation}. In \bibinfo{booktitle}{\emph{Proc. 20th ACM/IEEE Int. Conf. on Human-Robot Interaction (HRI)}}. IEEE, \bibinfo{pages}{1588--1592}.
\newblock
\href{https://doi.org/10.1109/HRI61500.2025.10974117}{doi:\nolinkurl{10.1109/HRI61500.2025.10974117}}


\bibitem[Serpiva et~al\mbox{.}(2025)]%
        {serpiva2025racevla}
\bibfield{author}{\bibinfo{person}{Valerii Serpiva}, \bibinfo{person}{Artem Lykov}, \bibinfo{person}{Artyom Myshlyaev}, \bibinfo{person}{Muhammad~Haris Khan}, \bibinfo{person}{Ali~Alridha Abdulkarim}, \bibinfo{person}{Oleg Sautenkov}, {and} \bibinfo{person}{Dzmitry Tsetserukou}.} \bibinfo{year}{2025}\natexlab{}.
\newblock \bibinfo{title}{RaceVLA: VLA-based Racing Drone Navigation with Human-like Behaviour}.
\newblock
\showeprint[arxiv]{2503.02572.}
\newblock
\shownote{Retrieved from https://arxiv.org/abs/2503.02572}.


\bibitem[Strabala et~al\mbox{.}(2013)]%
        {strabala2013toward}
\bibfield{author}{\bibinfo{person}{Kyle Strabala}, \bibinfo{person}{Min~Kyung Lee}, \bibinfo{person}{Anca Dragan}, \bibinfo{person}{Jodi Forlizzi}, \bibinfo{person}{Siddhartha~S Srinivasa}, \bibinfo{person}{Maya Cakmak}, {and} \bibinfo{person}{Vincenzo Micelli}.} \bibinfo{year}{2013}\natexlab{}.
\newblock \showarticletitle{Toward seamless human-robot handovers}.
\newblock \bibinfo{journal}{\emph{Journal of Human-Robot Interaction}} \bibinfo{volume}{2}, \bibinfo{number}{1} (\bibinfo{year}{2013}), \bibinfo{pages}{112--132}.
\newblock
\href{https://doi.org/10.5898/JHRI.2.1.Strabala}{doi:\nolinkurl{10.5898/JHRI.2.1.Strabala}}


\bibitem[Tsukagoshi et~al\mbox{.}(2015)]%
        {tsukagoshi2015aerial}
\bibfield{author}{\bibinfo{person}{H. Tsukagoshi}, \bibinfo{person}{M. Watanabe}, \bibinfo{person}{T. Hamada}, \bibinfo{person}{D. Ashlih}, {and} \bibinfo{person}{R. Iizuka}.} \bibinfo{year}{2015}\natexlab{}.
\newblock \showarticletitle{Aerial Manipulator with Perching and Door-Opening Capability}. In \bibinfo{booktitle}{\emph{Proc. IEEE Int. Conf. on Robotics and Automation (ICRA)}}. \bibinfo{publisher}{IEEE}, \bibinfo{pages}{4663--4668}.
\newblock
\href{https://doi.org/10.1109/ICRA.2015.7139845}{doi:\nolinkurl{10.1109/ICRA.2015.7139845}}


\bibitem[Wen et~al\mbox{.}(2025)]%
        {wen2025tinyvla}
\bibfield{author}{\bibinfo{person}{Junjie Wen}, \bibinfo{person}{Yichen Zhu}, \bibinfo{person}{Jinming Li}, \bibinfo{person}{Minjie Zhu}, \bibinfo{person}{Kun Wu}, \bibinfo{person}{Zhiyuan Xu}, \bibinfo{person}{Ning Liu}, \bibinfo{person}{Ran Cheng}, \bibinfo{person}{Chaomin Shen}, \bibinfo{person}{Yaxin Peng}, \bibinfo{person}{Feifei Feng}, {and} \bibinfo{person}{Jian Tang}.} \bibinfo{year}{2025}\natexlab{}.
\newblock \showarticletitle{TinyVLA: Toward Fast, Data-Efficient Vision-Language-Action Models for Robotic Manipulation}.
\newblock \bibinfo{journal}{\emph{IEEE Robotics and Automation Letters}} \bibinfo{volume}{10}, \bibinfo{number}{4} (\bibinfo{year}{2025}), \bibinfo{pages}{3988--3995}.
\newblock
\href{https://doi.org/10.1109/LRA.2025.3544909}{doi:\nolinkurl{10.1109/LRA.2025.3544909}}


\bibitem[Zhao et~al\mbox{.}(2022)]%
        {zhao2022forceful}
\bibfield{author}{\bibinfo{person}{M. Zhao}, \bibinfo{person}{K. Nagato}, \bibinfo{person}{K. Okada}, \bibinfo{person}{M. Inaba}, {and} \bibinfo{person}{M. Nakao}.} \bibinfo{year}{2022}\natexlab{}.
\newblock \showarticletitle{Forceful Valve Manipulation with Arbitrary Direction by Articulated Aerial Robot Equipped with Thrust Vectoring Apparatus}.
\newblock \bibinfo{journal}{\emph{IEEE Robotics and Automation Letters}} \bibinfo{volume}{7}, \bibinfo{number}{2} (\bibinfo{year}{2022}), \bibinfo{pages}{4893--4900}.
\newblock
\href{https://doi.org/10.1109/LRA.2022.3154018}{doi:\nolinkurl{10.1109/LRA.2022.3154018}}


\end{thebibliography}
\end{document}